\documentclass[10pt,twocolumn,letterpaper]{article}

\usepackage{iccv}
\usepackage{times}
\usepackage{epsfig}
\usepackage{graphicx}
\usepackage{amsmath}
\usepackage{amssymb}

\usepackage{subfigure}
\usepackage{tabularx}
\usepackage{color}
\usepackage{multirow}
\usepackage{rotating}
\usepackage{capt-of,etoolbox}
\usepackage{colortbl} 
\usepackage{arydshln}
\makeatletter
\renewcommand{\@thesubfigure}{\hskip\subfiglabelskip}
\makeatother
\usepackage{multirow}

\usepackage[breaklinks=true,bookmarks=false]{hyperref}

\iccvfinalcopy

\def\iccvPaperID{187}

\ificcvfinal\fi

\begin{document}

\title{GA-DAN: Geometry-Aware Domain Adaptation Network for Scene Text Detection and Recognition}

\author{Fangneng Zhan\\
Nanyang Technological University\\
50 Nanyang Avenue, Singapore 639798\\
{\tt\small fnzhan@ntu.edu.sg}
\and
Chuhui Xue\\
Nanyang Technological University\\
50 Nanyang Avenue, Singapore 639798\\
{\tt\small xuec0003@e.ntu.edu.sg}
\and
Shijian Lu\\
Nanyang Technological University\\
50 Nanyang Avenue, Singapore 639798\\
{\tt\small shijan.lu@ntu.edu.sg}
}

\maketitle
\ificcvfinal\thispagestyle{empty}\fi

\begin{abstract}
Recent adversarial learning research has achieved very impressive progress for modelling cross-domain data shifts in appearance space but its counterpart in modelling cross-domain shifts in geometry space lags far behind. This paper presents an innovative Geometry-Aware Domain Adaptation Network (GA-DAN) that is capable of modelling cross-domain shifts concurrently in both geometry space and appearance space and realistically converting images across domains with very different characteristics. In the proposed GA-DAN, a novel multi-modal spatial learning technique is designed which converts a source-domain image into multiple images of different spatial views as in the target domain. A new disentangled cycle-consistency loss is introduced which balances the cycle consistency in appearance and geometry spaces and improves the learning of the whole network greatly. The proposed GA-DAN has been evaluated for the classic scene text detection and recognition tasks, and experiments show that the domain-adapted images achieve superior scene text detection and recognition performance while applied to network training. 
\end{abstract}

\section{Introduction}

\begin{figure}[t]
\centering
\includegraphics[width=0.975\linewidth]{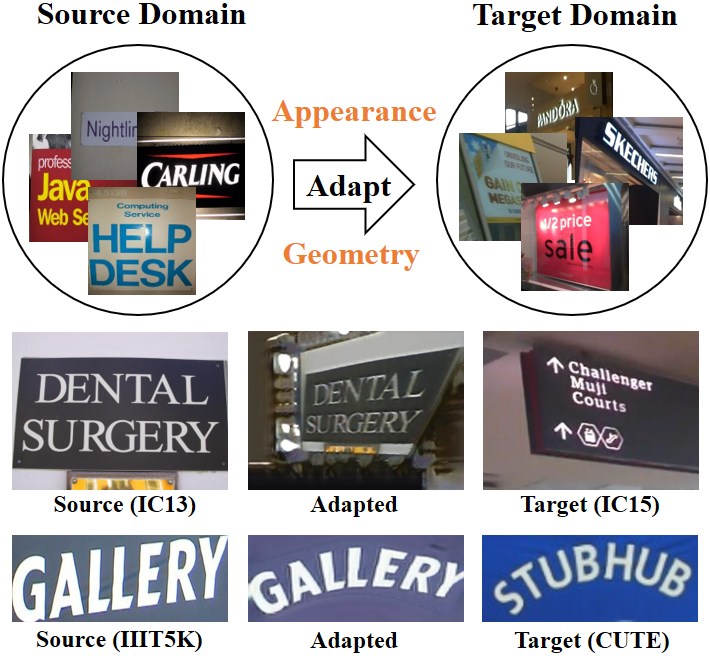}
\caption{Domain adaptation by the proposed GA-DAN: For scene text images with clear shifts from the \textbf{Source Domain} to the \textbf{Target Domain}, GA-DAN models the domain shifts in appearance and geometry spaces simultaneously and generates \textbf{Adapted images} with high-fidelity in both appearance and geometry spaces. 
}
\label{fig1}
\end{figure}

A large amount of labelled or annotated images is critical for training robust and accurate deep neural network (DNN) models, but collecting and annotating large datasets are often extremely expensive. 
In addition, state-of-the-art DNN models usually assume that images in the training and inference stages are collected under similar conditions which often experience clear performance drops while applied to images from different domains. Such lack of scalability and transferability makes collection and annotation even more expensive while dealing with images collected under different conditions from different domains. Unsupervised Domain Adaptation (DA), which transfers images and features from a source domain to a target domain, has achieved very impressive performance especially with the recent advances of Generative Adversarial Networks (GANs) \cite{goodfellow2014}. Different DA techniques have been developed and applied to different computer vision problems successfully such as style transfer, image synthesis, etc.

State-of-the-art DA still faces various problems. In particular, most existing systems focus on learning feature shift in appearance space only whereas the feature shift in geometry space is largely neglected. On the other hand, images from different domains often differ in both appearance and geometry spaces. Take various texts in scenes as an example. They could suffer from motion blurs in appearance space and perspective distortion in geometry space concurrently as shown in the target domain images in Fig. \ref{fig1}, and both are essential features for learning robust and accurate scene text detectors and recognizers. As a result, existing techniques often suffer from a clear performance drop when source and target domains have clear geometry discrepancy as observed for images and videos from different domains. 

We design an innovative Geometry-Aware Domain Adaptation Network (GA-DAN), an end-to-end trainable network that learns and models domain shifts in appearance and geometry spaces simultaneously as illustrated in the last two rows of Fig. \ref{fig1}. One unique feature of the proposed GA-DAN is a multi-modal spatial learning structure that learns multiple spatial transformations and converts a source-domain image into multiple target-domain images realistically as illustrated in Figs. 4 and 5. In addition, a novel disentangled cycle-consistency loss is designed which guides the GA-DAN learning towards optimal transfer concurrently in both geometry and appearance spaces. The GA-DAN takes the cycle structure as illustrated in Fig. \ref{fig2}, where the spatial modules ($S_X$ and $S_Y$) model the feature shifts in geometry space and the generators ($G_X$ and $G_Y$) complete the blank as introduced by the spatial transformation and model the feature shifts in appearance space. The discriminators discriminate not only `fake image' and `real image' but also `fake transformation' and `real transformation', leading to optimal modelling of domain and feature shifts in geometry and appearance spaces. 

The contributions of this work are threefold. First, it designs a novel network that models domain shifts in geometry and appearance spaces simultaneously. To the best of our knowledge, this is the first network that performs domain adaptation in both spaces concurrently. 
Second, it designs an innovative multi-modal spatial learning mechanism and introduces a spatial transformation discriminator to achieve multi-modal adaptation in geometry space. Third, it designs a disentangled cycle-consistency loss that balances the cycle-consistency for concurrent adaptation in appearance and geometry spaces and can also be applied to generic domain adaptation.

\section{Related Works}
\subsection{Domain Adaptation}

Domain adaptation is an emerging research topic that aims to address domain shift and dataset bias \cite{saenko2010, torralba2011}. Existing techniques can be broadly classified into two categories. The first category focuses on minimizing discrepancies between the source domain and the target domain in the feature space. For example, \cite{long2017} explored Maximum Mean Discrepancies (MMD) and Joint MMD distance across domains over fully-connected layers. \cite{sun_2016_2} studied feature adaptation by minimizing the correlation distance and then extended it to deep architectures \cite{sun2016}. \cite{bousmalis2016} modelled domain-specific features to encourage networks to learn domain-invariant features. \cite{ganin2015, tzeng2017} improved feature adaptation by designing various adversarial objectives.

The second category adopts Generative Adversarial Nets (GANs) \cite{goodfellow2014} to perform pixel-level adaptation via continuous adversarial learning between generators and discriminators which has achieved great success in image generation \cite{denton2015, radford2016, zhang2017}, image composition \cite{lin2018,zhan2019sfgan,zhan2019acgan} and image-to-image translation \cite{zhu2017, isola2017, shrivastava2017}. Different approaches have been investigated to address pixel-level image transfer by enforcing consistency in the embedding space. \cite{taigman2017} translates a rendering image to a real image by using conditional GANs. \cite{bousmalis2017} studies an unsupervised approach to learn pixel-level transfer across domains. \cite{liu2017unit} proposes an unsupervised image-to-image translation framework using a shared-latent space. \cite{dumoulin2016} introduces an inference model that jointly learns a generation network and an inference network.
More recently, CycleGAN \cite{zhu2017} and its variants \cite{yi2017, kim2017} achieve very impressive image translation by using cycle-consistency loss. \cite{hoffman2018} proposes a cycle-consistent adversarial model that adapts at both pixel and feature levels.

The proposed GA-DAN differs in two major aspects. First, GA-DAN adapts across domains in geometry and appearance spaces simultaneously while most existing works focus on pixel-level transfer in appearance space only. Second, the proposed disentangled cycle-consistency loss balances the cycle-consistency in both appearance and geometry spaces whereas existing works cannot. Note that \cite{lin2018} attempts to model geometry shifts very recently, but it focuses on geometry shifts in image composition only and also completely ignores appearance shifts.

\begin{figure*}
\centering
\includegraphics[width=0.95\linewidth]{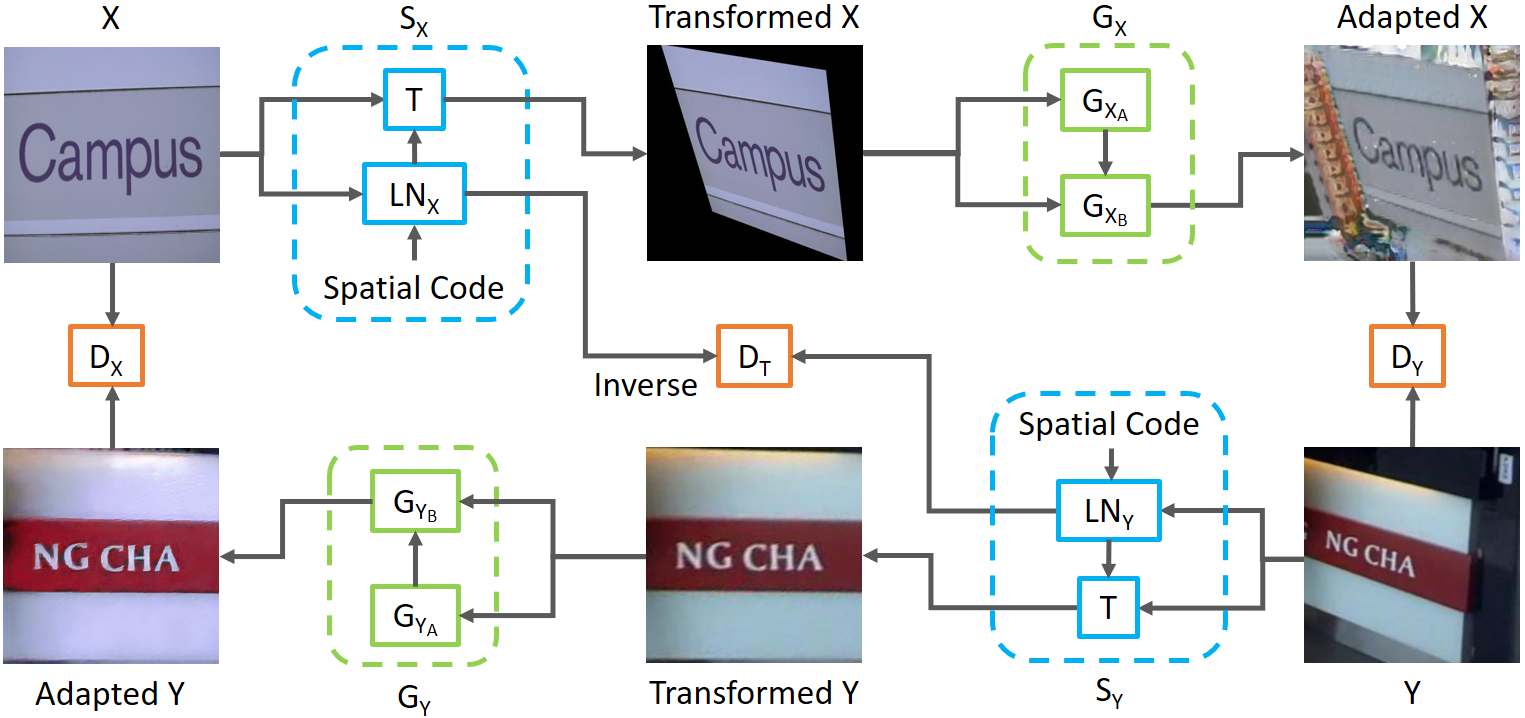}
\caption{The structure of the proposed GA-DAN: $S_{X}$ (or $S_{Y}$) represents the spatial modules as enclosed in blue-color boxes which consist of \textit{Spatial Code}, transformation module $T$ and localization network $LN_{X}$ (or $LN_{Y}$) that predict transformation matrix and transform input images. $G_{X}$ (or $G_{Y}$) denote generators consisting of $G_{X_A}$ (or $G_{Y_A}$) and $G_{X_B}$ (or $G_{Y_B}$) as enclosed in green-color boxes that complete the background and translate the image style, respectively. $D_{X}$, $D_{Y}$ and $D_{T}$ within orange-color boxes denote different discriminators.}
\label{fig2}
\end{figure*}

\subsection{Scene Text Detection and Recognition}
Automated detection and recognition of various texts in scenes has attracted increasing interests as witnessed by increasing benchmarking competitions \cite{icdar2015,icdar2017}. Different detection techniques have been proposed from those earlier using hand-crafted features \cite{neumann2013,lu2015} to the recent using DNNs \cite{zhang2018,jaderberg2016,yin2015,tian2016,zhan2018ver,xue2018}. Different detection approaches have also been explored including character based \cite{huang2014,tian2015,jaderberg2014,he2016,Hu2017ICCVWordSup}, word-based \cite{jaderberg2016,liao2017,liu2017_2,he2017,east,liu2018fots,Wang_2018_CVPR,Lyu_2018_ECCV,lyu2018multi,polzounov2017wordfence,zhan2019scene,deng2018pixellink,long2018textsnake,liao2018rotation} and the recent line-based \cite{zhang2015_2}. Meanwhile, different scene text recognition techniques have been developed from the earlier recognizing characters directly \cite{jaderberg2014_2,yao2014,rodrguez2015,almazan2014,gordo2015,jaderberg2015} to the recent recognizing words or text lines using recurrent neural network (RNN), \cite{shi2017,su2014,su2017,shi2016} and attention models \cite{lee2016,cheng2017,zhan2019esir}.

Similar to other detection and recognition tasks, training accurate and robust scene text detectors and recognizers requires a large amount of annotated training images. On the other hand, most existing datasets such as ICDAR2015 \cite{icdar2015} and Total-Text \cite{chng2017} contain a few hundred or thousand training images only which has become one major factor that impedes the advance of scene text detection and recognition research. The proposed domain adaptation technique addresses this challenge by transferring existing annotated scene text images to a new target domain, hence alleviate the image collection and annotation efforts greatly.

\section{Methodology}
We propose an innovative geometry-aware domain adaptation network (GA-DAN) that performs multi-modal domain adaptation concurrently in both spatial and appearance spaces as shown in Fig. \ref{fig2}. Detailed network architecture, multi-modal spatial learning and adversarial training strategy will be presented in the following three subsections.


\subsection{GA-DAN Architecture} \label{sec:net}
The GA-DAN consists of spatial modules, generators and discriminators as enclosed within blue-color, green-color and orange-color boxes, respectively, as shown in Fig. \ref{fig2}. The overall network is designed in a cycle structure, where the mappings between the source domain $X$ and the target domain $Y$ are learned by sub-modules $X \rightarrow Adapted \ X$ ($X \rightarrow Y$) and $Y \rightarrow Adapted \ Y$ ($Y \rightarrow X$), respectively. In the $X \rightarrow Y$ mapping, the spatial module $S_{X}$ transforms images in $X$ to new images in \textit{Transformed X} that has similar spatial styles as $Y$. The generator $G_{X}$ then completes the blank as introduced by the spatial transformation and translates the completed images to new images in \textit{Adapted X} that has similar appearance as $Y$. A discriminator $D_{Y}$ attempts to distinguish \textit{Adapted X} and $Y$ which drives $S_{X}$ and $G_{X}$ to learn better spatial and appearance mapping from $X$ to $Y$. Similar processes happen in the $Y\rightarrow X$ mapping as well.

The spatial modules $S_{X}$ (as well as $S_{Y}$) has a localization network $LN_{X}$ and a transformation module $T$ for domain adaptation in geometry space, more details to be presented in the following subsection. The generator $G_{X}$ (as well as $G_{Y}$) consists of two sub-generators $G_{X_{A}}$ and $G_{X_{B}}$ for adaptation in appearance space. In particular, the spatial module $S_X$ will produce a binary map with 1 denoting pixels transformed from the original image and 0 for padded black background (not shown but can be inferred from the sample image in Transformed X in Fig. \ref{fig2}). Under the guidance of the binary map, $G_{X_A}$ will learn from $Y$ for new contents to complete the black background of the transformed image (as in \textit{Transformed X}), and $G_{X_B}$ further adapts the completed images to have similar appearance as $Y$ as illustrated in Fig. \ref{fig2}. Our study shows that the \textit{Adapted X} is quite blurry if a single generator is used to complete the black background and adapt the appearance. The use of the two dedicated generators $G_{X_A}$ and $G_{X_B}$ for background completing and appearance adaptation helps greatly for realistic adaptation in appearance space.

Note directly concatenating an appearance-transfer GAN (e.g. CycleGAN \cite{zhu2017}) and a geometry-transfer GAN (e.g. ST-GAN \cite{lin2018}) does not perform well for simultaneous image adaptation in geometry and appearance spaces. Due to the co-existence of spatial and appearance shifts between the source and target domain images, the discriminator of the geometry-transfer GAN (or appearance-transfer GAN) will be confused by the appearance (or geometry) shift which leads to poor adversarial learning outcome. Our GA-DAN is an end-to-end trainable network that coordinates the learning in geometry and appearance spaces simultaneously and drives the network for optimal adaptation in both spaces, more details to be presented in Section \ref{sec:loss}.

\begin{table}[ht]
\caption{Localization network $LN_{X}$ and $LN_{Y}$ within the multi-modal spatial learning shown in Fig. \ref{fig2}, N denotes the number of parameters.}
\centering 
\begin{tabular}{|c|c|c|} 
\hline 
\textbf{Layers} & \multicolumn{1}{c|}{\textbf{Out Size}} & \multicolumn{1}{c|}{\textbf{Configurations}} \\
\hline
Block1 & $128 \times 128$ & $3 \times 3 \ conv, 16, 2 \times 2 \ pool$ \\
\hline
Block2 & $64 \times 64$ & $3 \times 3 \ conv, 32, 2 \times 2 \ pool$ \\
\hline
Block3 & $32 \times 32$ & $3 \times 3 \ conv, 64, 2 \times 2 \ pool$ \\
\hline
Block4 & $16 \times 16$ & $3 \times 3 \ conv, 128, 2 \times 2 \ pool$ \\
\hline
Block5 & $8 \times 8$ & $3 \times 3 \ conv, 128, 2 \times 2 \ pool$ \\
\hline
FC1 & $512$ & - \\
\hline
FC2 & N & - \\
\hline
\end{tabular}
\label{tab:stn}
\end{table}

\subsection{Multi-Modal Spatial Learning}
To generate images with different spatial views and features (similar to images in the target domain), we design a multi-modal spatial learning structure that learns multi-modal spatial transformations and maps a source-domain image to multiple target-domain images with different spatial views. Specifically, the multi-modal spatial learning first samples \textit{Spatial Code} (i.e., random vectors) from normal distributions and then regresses it to predict spatial transformation matrix (according to the spatial features of the input image) by using a localization network $LN_{X}$ (or $LN_{Y}$) as shown in Table \ref{tab:stn}. With the predicted transformation matrix that could be affine, homography or thin plate spline \cite{tps}, the input image can be transformed to a new image with a different spatial view by $T$ which performs actual transformation. Multiple new spatial views can be generated by running GA-DAN and sampling the \textit{Spatial Code} multiple times, leading to the proposed multi-modal spatial mapping as illustrated in Figs. \ref{fig:det} and \ref{fig:det2}. 

The multi-modal spatial learning as guided by $D_{X}$ and $D_{Y}$ tends to be unstable and hard to converge as the concurrent learning in geometry and appearance spaces is over-flexible and often entangled with each other. We address this issue by including a new discriminator $D_{T}$ as shown in Fig. \ref{fig2} which imposes certain constraints to the cyclic spatial learning and accordingly leads to more stable and efficient learning of the whole network. As shown in Fig. \ref{fig2}, $S_{X}$ predicts a transformation matrix $H_{XY}$ for mapping from domain $X$ to domain $Y$, and $S_{Y}$ predicts another transformation matrix $H_{YX}$ for mapping from domain $Y$ to domain $X$. The inverse matrix $H_{XY}^{-1}$ can be derived from $H_{XY}$ and it should be in the same transformation domain with the $H_{YX}$. $D_{T}$ thus attempts to discriminates $H_{XY}^{-1}$ and $H_{YX}$ which drives the spatial transformations in two inverse directions to learn from each other. It bridges the spatial learning in opposite directions and imposes extra constraints in the geometry spaces, greatly improving the learning efficiency and learning stability of the whole network.

\subsection{Adversarial Training} \label{sec:loss}
Due to the adaptation in geometry and appearance spaces, the adversarial learning needs to coordinate the minimization of cycle-consistency loss in both spaces properly. In addition, the adversarial learning also needs to take care of the new discriminator $D_T$ as shown in Fig. \ref{fig2}. We design an innovative disentangled cycle-consistency loss and adversarial objective to tackle these challenges, more details to be described in the following two subsections.

\begin{figure}[t]
\centering
\includegraphics[width=0.95\linewidth]{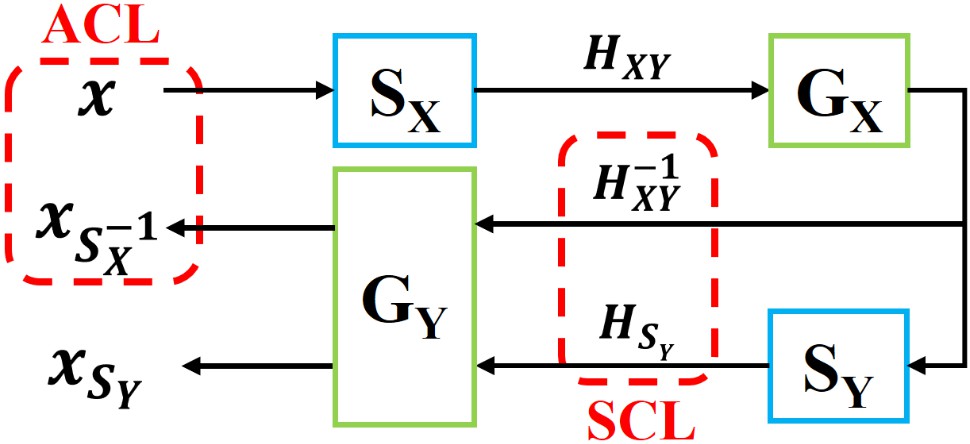}
\caption{Illustration of the disentangled cycle-consistency loss: $S_{X}$, $S_{Y}$, $G_{X}$ and $G_{Y}$ denotes the spatial modules and generators, respectively, as shown in Fig. \ref{fig2}. $x$, $x_{S_{X}^{-1}}$ and $x_{S_{Y}}$ refer to the input images in domain X, reconstructed image by the inverse transformation of $S_{X}$ and reconstructed image by $S_{Y}$. $H_{XY}$ and $H_{S_{Y}}$ refer to the predicted transformation matrices by $S_X$ and $S_Y$. ACL and SCL denote appearance cycle-consistency loss and spatial cycle-consistency loss which are obtained by calculating the L1 loss between ($x,x_{S_{X}^{-1}}$) and ($H_{XY}^{-1}, H_{S_{Y}}$), respectively.}
\label{fig:loss}
\end{figure}

\medskip \noindent 
\textbf{Disentangled Cycle-Consistency Loss.} We design a disentangled cycle-consistency loss that decomposes the cycle-consistency loss into a spatial cycle-consistency loss (SCL) and an appearance cycle-consistency loss (ACL) and balances their weights during learning. With spatial transformation involved, a small shift (due to inaccurate prediction of the spatial transformation matrix) in geometry space will lead to a very large cycle-consistency loss which can easily override the ACL and ruin the learning of the whole network. The decomposition of the cycle-consistency loss into ACL and SCL helps to address this issue effectively.

As shown in Fig. \ref{fig:loss}, the image $x$ is fed into $S_{X}$ to predict the transformation matrix $H_{XY}$ and the transformed image is then fed to $G_{X}$ for translation in appearance space. The translated image will be recovered in two different manners. First, it will be transformed by the inverse of $H_{XY}$ (i.e. $H_{XY}^{-1}$) and further translated by $G_{Y}$ to generate $x_{S_{X}^{-1}}$. Second, it will be passed to $S_{Y}$ to predict the transformation matrix $H_{S_{Y}}$ so as to be transformed by the estimated $H_{S_{Y}}$ and further translated by $G_{Y}$ to produce $x_{S_{Y}}$. Note the \textit{Spatial Code} in $S_{X}$ and $S_{Y}$ are identical here so that the input image can be recovered in geometry space.

The $x_{S_{X}^{-1}}$ can be fully recovered from $x$ in geometry space since the recovering matrix $H_{XY}^{-1}$ is simply the inverse of $H_{XY}$. But $x_{S_{X}^{-1}}$ is different from $x$ in appearance space. The ACL can thus be computed by L1 loss between $x$ and $x_{S_{X}^{-1}}$ (only appearance difference exists) as follows:
\begin{equation}
\begin{split}
ACL_{X} = E_{x \sim X}[\left \| x_{S_{X}^{-1}} - x \right \|]
\end{split}
\end{equation}
Though $x_{S_{X}^{-1}}$ and $x_{S_{Y}}$ differ only in geometry space, SCL cannot be obtained by computing L1 loss between them because a minor shift in geometry space will lead to a very large L1 loss. To ensure the spatial cycle-consistency, we obtain the SCL by directly computing the L1 loss of the transformation matrix $H_{XY}^{-1}$ and $H_{S_{Y}}$ as follows:
\begin{equation}
\begin{split}
SCL = E_{x \sim X}[\left \| H_{XY}^{-1} - H_{S_{Y}} \right \|]
\end{split}
\end{equation}
Further, the bordering regions of the original image may be lost by the spatial transformation which could affect the training seriously. While adapting an image from domain $X$ to domain $Y$, the adaptation should ensure that all image information within the domain $X$ is well preserved. Given the binary transformation map $m$ from $S_{X}$, we can directly apply the inverse transformation $H_{XY}^{-1}$ to $m$ to obtain $m_{H_{XY}^{-1}}$. As the missing region by the spatial transformation will not be recovered, a region missing loss (RML) is defined for better preserving the transformed image as follows:
\begin{equation}
\begin{split}
RML = E_{x \sim X}[\left \| m_{H_{XY}^{-1}} - m \right \|]
\end{split}
\end{equation}
The overall cycle-consistency loss in the domain $X$ can thus be formed as follows:
\begin{equation}
\begin{split}
L_{cyc} = \lambda_{acl} ACL + \lambda_{scl} SCL + RML
\end{split}
\end{equation}
where $\lambda_{acl}$ and $\lambda_{scl}$ are the weights of $ACL$ and $SCL$.


\medskip \noindent 
\textbf{Adversarial Objective.} 
The adversarial objective of the mapping $X \rightarrow Y$ can be defined by:
\begin{equation}
\begin{split}
& L_{GAN} = E_{y \sim Y }[\log D_{Y}(y)] \\
& + E_{x \sim X }[\log (1 - D_{Y}(G_{X}(S_{X}(x)))] \\
& + E_{y \sim Y }[\log D_{T}(H_{YX}^{-1})] + E_{x \sim X }[\log (1- D_{T}(H_{XY}))]
\end{split}
\end{equation}
where $H_{XY}$ and $H_{YX}^{-1}$ are the transformation matrix for $X \rightarrow Y$ and the inverse transformation for $Y \rightarrow X$. $S_{X}$ and $G_{X}$ aim to minimize this objective while $D_{Y}$ and $D_{T}$ try to maximize it, i.e. $minS_{X}G_{X}maxD_{Y}D_{T} L_{GAN}$.
The objective of the mapping $Y \rightarrow X$ can be obtained similarly. Note to ensure that the translated image preserves features of the original image, an identity loss is included as follows: 
\begin{equation}
L_{idt} = E_{x \sim X} [\left \| G_{X}(S_{X}(x))*m - S_{X}(x)*m \right \|]
\end{equation}
where $m$ refers to the binary mask as produced by $S_X$.

\renewcommand\arraystretch{1.5}
\begin{table*}[t]
\centering
\caption{Scene text detection over the test images of the target datasets ICDAR2015 and MSRA-TD500: `IC13', `Target', `AD-IC13' and `10-AD-IC13' denote the dataset ICDAR2013, target dataset (ICDAR2015 or MSRA-TD500), 1-to-1 adapted ICDAR2013 and 1-to-10 adapted ICDAR2013, respectively. `SynthText' refers to 800K synthetic images as reported in \cite{gupta2016synthtext}.}
\begin{tabular}{|l|c|c|c|c|c|c|}
\hline
                                                        & \multicolumn{3}{c|}{\textbf{ICDAR2015}} & \multicolumn{3}{c|}{\textbf{MSRA-TD500}} \\ \hline
\textbf{Method}                                         & \textbf{Recall}  & \textbf{Precision}  & \textbf{F-score}  & \textbf{Recall}   & \textbf{Precision}  & \textbf{F-score}  \\ \hline \hline
RRD \cite{liao2018rotation} [SynthText + Target]        & 79.0        & 85.6        & 82.2        & 73.0         & \textbf{87.0}        & \textbf{79.0} \\ \hline
TextSnake \cite{long2018textsnake} [SynthText + Target] & 80.4        & 84.9        & 82.6        & \textbf{73.9}         & 83.2        & 78.3        \\ \hline
\hline
EAST [IC13]                                     & 43.7        & 68.2        & 53.3        & 34.9         &  71.2       & 46.8        \\ \hline
EAST [AD-IC13]                                  & 59.6        & 69.9        & 64.4        & 51.5         &  67.7       & 58.5        \\ \hline
EAST [10-AD-IC13]                               & 71.6        & 67.3        & 69.4        & 55.8         &  69.9       & 62.1        \\ \hline\hline
EAST [Target]                                   & 76.9        & 81.1        & 79.0        & 64.4     & 73.8     & 68.7    \\ \hline
EAST [IC13 + Target]                            & 77.0        & 83.2        & 80.0        & 66.2         &  74.8       & 70.3        \\ \hline
EAST [AD-IC13 + Target]                         & 79.2        & 83.7        & 81.4        & 67.7         &  77.5       & 72.3        \\ \hline
EAST [10-AD-IC13 + Target]                      & \textbf{81.6}        & \textbf{85.6}        & \textbf{83.5}        & 71.1         &  80.5       & 75.5        \\ \hline
\end{tabular}
\label{tab:det_data}
\end{table*}
\section{Experiments}
The proposed image adaptation technique has been evaluated over the scene text detection and recognition tasks.
\subsection{Datasets}
The experiments involve seven publicly available scene text detection and recognition datasets as listed:

\textbf{ICDAR2013} \cite{icdar2013} is used in the Robust Reading Competition in the International Conference on Document Analysis and Recognition (ICDAR) 2013. The images explicitly focused around the text content of interest. It contains 848 word images for network training and 1095 for testing.

\textbf{ICDAR2015} \cite{icdar2015} is used in the Robust Reading Competition under ICDAR2015. It contains incidental scene text images that appears in the scene without taking any specific prior action to improve its positioning / quality in the frame.
2077 text image patches are cropped from this dataset, where a large amount of cropped scene texts suffer from perspective and curvature distortions.

\textbf{MSRA-TD500} \cite{msra-td500} dataset consists of 500 natural images (300 for training, 200 for test), which are taken from indoor and outdoor scenes using a pocket camera. The indoor images mainly capture signs, doorplates and caution plates while the outdoor images mostly capture guide boards and billboards with complex background.

\textbf{IIIT5K} \cite{iiit5k} has 2000 training images and 3000 test images that are cropped from scene texts and born-digital images. Each word in this dataset has a 50-word lexicon and a 1000-word lexicon, where each lexicon consists of a ground-truth word and a set of randomly picked words.

\textbf{SVT} \cite{wang2011} is collected from the Google Street View images that were used for scene text detection research. 647 words images are cropped from 249 street view images and most cropped texts are almost horizontal.

\textbf{SVTP} \cite{phan2013} has 639 word images that are cropped from the SVT images. Most images in this dataset suffer from perspective distortion which are purposely selected for evaluation of scene text recognition under perspective views.

\textbf{CUTE} \cite{risnumawan2014} has 288 word images most of which are curved. All words are cropped from the CUTE dataset which contains 80 scene text images that are originally collected for scene text detection research.

\subsection{Scene Text Detection}

The proposed GA-DAN is evaluated by the performance of the scene text detectors that are trained by using its adapted images. In evaluations, the training set of ICDAR2013 (IC13) is used as the source dataset and the training sets of ICDAR2015 (IC15) and MSRA-TD500 (MT) are used as the target datasets which contain very different images as compared with those in IC13. GA-DAN generates two sets of images `AD-IC13' and `10-AD-IC13' for each of the two target datasets. The `AD-IC13' is generated by 1-to-1 adaptation where each IC13 image is transformed to a single image that has similar geometry and appearance as the target dataset. The `10-AD-IC13' is produced by 1-to-10 adaptation where each IC13 image is transformed to 10 adapted images by 
sampling 10 different spatial codes. Scene text detector EAST \cite{east} is adopted for evaluation.

Table \ref{tab:det_data} shows quantitative results on the test set of two target datasets. Seven EAST models are trained for each target dataset by using different training images including 
1) [IC13]: the training set of IC13, 2) [AD-IC13]: the 1-to-1 adapted IC13, 3) [10-AD-IC13]: the 1-to-10 adapted IC13, 4) [Target]: the training set of each target dataset, 5) [IC13 + Target]: the combination of the IC13 training set and the training set of each target dataset, 6) [AD-IC13 + Target]: the combination of AD-IC13 and the training set of each target dataset, and 7) [10-AD-IC13 + Target]: the combination of 10-AD-IC13 and the training set of each target dataset.

As Table \ref{tab:det_data} shows, the effectiveness of GA-DAN adapted images can be observed from three aspects. First, \textbf{EAST [AD-IC13]} outperforms \textbf{EAST [IC13]} by f-scores of 11.1\% ($53.3\% \rightarrow 64.4\%$) and 11.7\% ($46.8\%\rightarrow58.5\%$) on the target datasets IC15 and MT, respectively, demonstrating the effectiveness of GA-DAN in adapting images from IC13 to IC15 and MT. Second, \textbf{EAST [10-AD-IC13]} improves \textbf{EAST [AD-IC13]} by f-scores of 5\% ($64.4\% \rightarrow 69.4\%$) and 3.6\% ($58.5\% \rightarrow 62.1\%$) on IC15 and MT, respectively. This shows the effectiveness of the multi-modal spatial learning that transforms a source-domain image to multiple target-domain images that are complementary with different spatial views. Third, \textbf{EAST [10-AD-IC13+Target]} improves \textbf{EAST [IC13+Target]} by a f-score of 3.5\% ($80.0\% \rightarrow 83.5\%$) and 5.2\% ($70.3\% \rightarrow 75.5\%$) on IC15 and MT, respectively. This shows that the adapted images are clearly more useful when combined with the training images of the target datasets for model training. 
\begin{figure*}[ht]
\centering
\subfigure {\includegraphics[width=.98\linewidth]{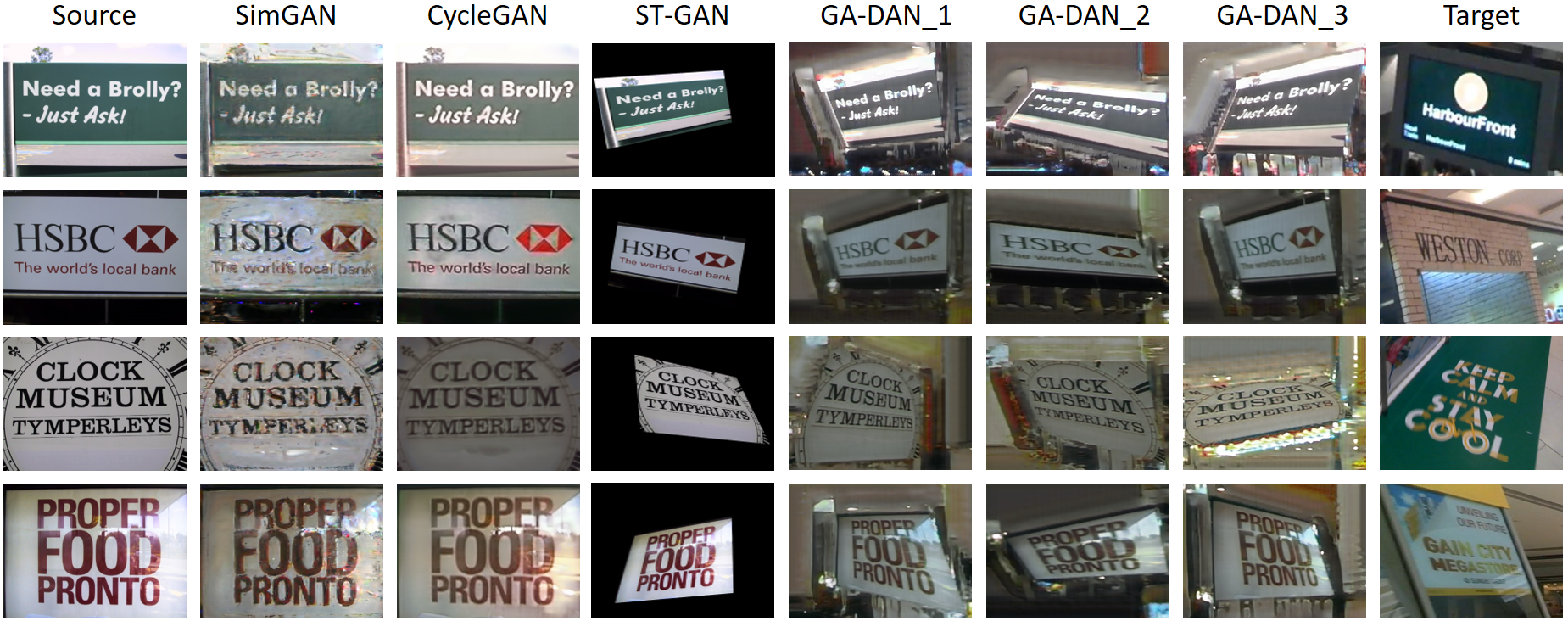}}
\vspace{0.01 pt}
\caption{Comparing our GA-DAN with state-of-the-art adaptation methods: The first and last columns show source-domain (IC13) and target-domain (IC15) images. GA-DAN\_1, GA-DAN\_2 and GA-DAN\_3 show three GA-DAN adapted images of different spatial views.}
\label{fig:det}
\end{figure*}

In addition, \textbf{EAST [10-AD-IC13+Target]} achieves state-of-the-art performance on the dataset IC15 by including only 2.3K GA-DAN adapted images (from 230 IC13 training images). As a comparison, TextSnake and RRD use 800K synthetic images in `SynthText' \cite{jaderberg2014_2} and they are also more advanced scene text detectors. Though the `10-AD-IC13' is much smaller than SynthText, it contributes more to the detection improvement largely because of the large domain shifts between SynthText and IC15. For the target dataset MT, the f-score of \textbf{EAST [10-AD-IC13+Target]} is slightly lower than that of state-of-the-art detectors TextSnake and RRD, largely because the domain shifts between MT and SynthText are relatively small and the much larger amount of images in SynthText help more on the performance improvement. We believe higher f-score can be achieved when a higher number of GA-DAN adapted images are included in model training. 

Table \ref{tab:det_model} shows the detection performance of different domain adaptation methods when EAST are trained by using their adapted images from IC13 to IC15 (the Baseline is trained using the original IC13 training images). Note for CycleGAN we adopt patch-wise training to minimize the effect of geometry differences in adversarial training. As ST-GAN is originally for image composition, we adapt it to achieve image translation in geometry space and restrict the transformation parameters to avoid boundary losing in testing phase. As Table \ref{tab:det_model} shows, all three adaptation models GA-DAN, CycleGAN and ST-GAN outperform Baseline clearly, and GA-DAN achieves clearly better f-score (64.4\% vs. 57.2\% and 57.6\%), demonstrating its superiority in adapting more realistic images. We also evaluate a new model ST-GAN + CycleGAN that directly concatenates ST-GAN and CycleGAN for adaptation in both geometry and appearance spaces. It shows that our GA-DAN still performs better by a large margin in f-score (64.4\% vs. 60.8\%), demonstrating its advantages in concurrent learning of geometric and appearance features.

Fig. \ref{fig:det} compares our GA-DAN with several state-of-the-art image adaptation methods. As Fig. \ref{fig:det} shows, GA-DAN adapts in both appearance and geometric spaces realistically, whereas SimGAN and CycleGAN can only adapt appearance features and ST-GAN can only adapt geometric features. In addition, GA-DAN\_1, GA-DAN\_2 and GA-DAN\_3 show three GA-DAN adapted images with different spatial views, demonstrating the effectiveness of our proposed multi-modal spatial learning.

\begin{table}[ht]
\caption{Scene text detection on the IC15 test images: The detection models are trained using the adapted IC13 training images (from IC13 to IC15) by different adaptation methods as listed. (Baseline is trained by using the original IC13 training images)
}
\centering
\begin{tabular}{|c|c|c|c|}
\hline
\textbf{Method}           & \textbf{Recall} & \textbf{Precision} & \textbf{F-score} \\ \hline \hline
CycleGAN \cite{zhu2017}  & 50.3       & 66.3       & 57.2   \\ \hline
ST-GAN \cite{lin2018}     & 52.9       & 63.4       & 57.6   \\ \hline
ST-GAN + CycleGAN           & 57.3       & 64.7       & 60.8   \\ \hline \hline
Baseline                  & 43.7       & 68.2       & 53.3       \\ \hline
GA-DAN                    & \textbf{59.6}       & \textbf{69.9}       & \textbf{64.4}       \\ \hline
\end{tabular}
\label{tab:det_model}
\end{table}

\subsection{Scene Text Recognition}

\begin{figure*}[t]
\centering
\subfigure {\includegraphics[width=.98\linewidth]{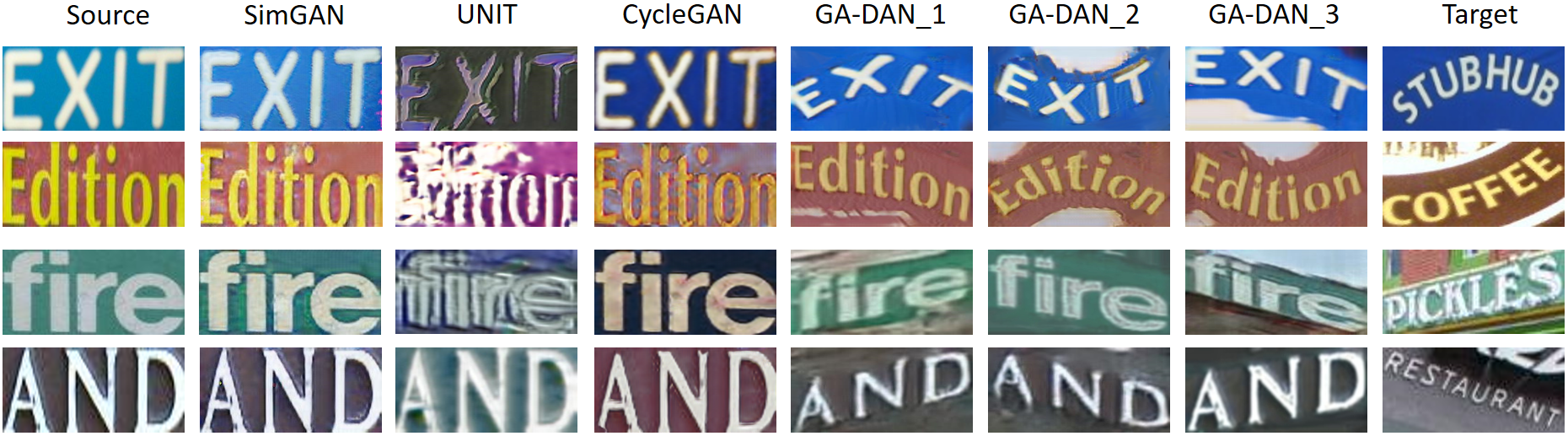}}
\vspace{0.01 pt}
\caption{Comparing GA-DAN with state-of-the-art adaptation methods: Rows 1-2 show adaptation from COMB to CUTE, Rows 3-4 show adaptation from COMB to SVTP. GA-DAN\_1, GA-DAN\_2 and GA-DAN\_3 show GA-DAN adapted images of different spatial views.}
\label{fig:det2}
\end{figure*}

For scene text recognition experiment, we select the CUTE \cite{risnumawan2014} and SVTP \cite{phan2013} as the target datasets.
As current scene text recognition datasets are all too small, we combine all images from datasets IC13 \cite{icdar2013}, IIIT5K \cite{iiit5k} and SVT \cite{wang2011} as the source dataset denoted by `COMB'. As scene texts in CUTE and SVTP are most curved or in perspective views but most COMB texts are horizontal, we use the thin plate spline for spatial transformation which is flexible for various spatial transformations. 

\renewcommand\arraystretch{1.75}
\begin{table}[ht]
\caption{Ablation study and comparisons with state-of-the-art adaptation methods: Recognition models are trained by different adaptations of the COMB images to the target domains CUTE and SVTP (Baseline uses the original COMB images and `Random' applies random spatial transformation in adaptation).}
\centering 
\begin{tabular}{|p{2.5cm} | p{1.75cm}<{\centering} | p{1.75cm}<{\centering} |}
\hline
\bfseries Methods & \multicolumn{1}{c|}{\bfseries COMB-CUTE} & \multicolumn{1}{c|}{\bfseries COMB-SVTP} \\\hline
SimGAN \cite{liu2016} & 30.7 & 42.6 \\\hline
UNIT \cite{liu2017unit} & 28.7 & 40.8 \\\hline
CoGAN \cite{yi2017} & 28.3 & 40.2 \\\hline
DualGAN \cite{liu2016} & 31.5 & 42.7 \\\hline
CycleGAN \cite{zhu2017} & 31.9 & 43.0 \\\hline
CyCADA \cite{hoffman2018} & 32.2 & 43.6 \\\hline
\hline
Baseline & 30.9 & 42.5 \\\hline
Random & 28.8 & 42.7 \\\hline
GA-DAN \ [WD] & 32.6 & 44.9 \\\hline
GA-DAN \ [WA] & 36.1  & 45.2 \\\hline
GA-DAN \ [WM] & 38.2  & 47.1 \\\hline
GA-DAN \ [10 \ AD] & 43.1 & 51.7 \\\hline
\end{tabular}
\label{tab:PPer}
\end{table}

Table \ref{tab:PPer} shows recognition accuracy when COMB images are adapted by different adaptation methods and then used to train the scene text recognition model: MORAN \cite{luo2019}. As Table \ref{tab:PPer} shows, \textbf{GA-DAN [WM]} (GA-DAN with 1-to-1 spatial transformation) outperforms other adaptation methods with a large margin. Additionally, most compared adaptation methods do not show clear improvement over the Baseline (trained by using the original COMB images without adaptation). In particular, CycleGAN and CyCADA improve the accuracy by 1.0\% and 1.3\% only for CUTE because they only adapt in appearance space but the main discrepancy between COMB and CUTE is in geometry space. CoGAN and UNIT tend to over-adapt the text appearance which may even change the text semantics and make texts unrecognizable. 

Table \ref{tab:PPer} also shows the ablation study results. Two GA-DAN models are trained for image adaptation. The first model is a complete GA-DAN with all newly designed features and components includes. The second is \textbf{GA-DAN [WD]} which is trained with a normal instead of disentangled cycle-consistency loss. 
For fair comparison, the region missing loss is also included in \textbf{GA-DAN [WD]}. For the complete GA-DAN, three sets of adapted images are generated to train the recognition model. The first set is \textbf{GA-DAN [WA]} that just takes the output of $S_{X}$ without appearance adaptation as shown in Fig. \ref{fig2}. The second set is \textbf{GA-DAN [WM]} that performs 1-to-1 adaptation and transforms each source-domain image into a single target-domain image. The third set is \textbf{GA-DAN [10 AD]} that performs 1-to-10 adaptation and transforms each source-domain image into 10 target-domain images. As Table \ref{tab:PPer} shows, \textbf{GA-DAN [WA]} clearly outperforms Baseline and `Random' (adapted using a random spatial transformation matrix) as well as state-of-the-art adaptation methods, showing the superiority of our spatial module in learning correct and accurate spatial transformations. \textbf{GA-DAN [WD]} outperforms state-of-the-art methods but clearly performs worse than \textbf{GA-DAN [WM]}, demonstrating the effectiveness of the proposed disentangled cycle-consistency loss. \textbf{GA-DAN [10 AD]} outperforms \textbf{GA-DAN [WM]} clearly, demonstrating the effectiveness of our proposed multi-modal spatial learning. 

Fig. \ref{fig:det2} compares our GA-DAN with several state-of-the-art adaptation methods. As Fig. \ref{fig:det2} shows, GA-DAN adapts in both appearance and geometry spaces realistically whereas CycleGAN and SimGAN can only adapt in appearance space. 
In addition, GA-DAN\_1, GA-DAN\_2 and GA-DAN\_3 show that the proposed GA-DAN is capable of transforming a source-domain image to multiple target-domain images of different spatial views.

\section{Conclusions}
This paper presents a geometry-aware domain adaptation network that achieves domain adaptation in geometry and appearance spaces simultaneously. A multi-modal spatial learning technique is proposed which can generate multiple adapted images with different spatial views. A novel disentangle cycle-consistency loss is designed which greatly improves the stability and concurrent learning in both geometry and appearance spaces. The proposed network has been validated over scene text detection and recognition tasks and experiments show the superiority of the adapted images while applied to train deep networks. 

{\small
\bibliographystyle{ieee}
\bibliography{egbib}
}

\end{document}